\documentclass[letterpaper, 10 pt, conference]{styles/ieeeconf}  

\IEEEoverridecommandlockouts                              

\usepackage{cite}
\usepackage{caption} 
\usepackage{times}

\usepackage{epsfig}
\usepackage{graphicx}
\usepackage{amsmath}
\usepackage{amssymb}

\usepackage{tabularx}

\usepackage[utf8]{inputenc}
\usepackage[T1]{fontenc}
\usepackage{textcomp}
\usepackage{gensymb}
\usepackage{multirow}
\usepackage{booktabs}
\usepackage{xcolor}
\usepackage{graphicx}
\usepackage{subcaption}
\usepackage{float}
\usepackage{capt-of}
\usepackage{etoolbox}
\usepackage{adjustbox}
\usepackage[font=small]{caption}
\usepackage{hyperref}

\title{\textbf{Simple means Faster: Real-Time Human Motion Forecasting in Monocular First Person Videos on CPU}}

\author{
Junaid Ahmed Ansari and Brojeshwar Bhowmick \\
\small{TCS Research and Innovation Labs}\\
 \small{Kolkata, India}\\
 \small{\{junaidahmed.ansari, b.bhowmick\}@tcs.com}
}

\begin{document}

\bstctlcite{Force_Etal}

\maketitle

\addtocounter{figure}{0}



\begin{abstract}
We present a simple, fast, and light-weight RNN based framework for forecasting future locations of humans in first person monocular videos. The primary motivation for this work was to design a network which could accurately predict future trajectories at a very high rate on a CPU. Typical applications of such a system would be a social robot or a visual assistance system ``for all'', as both cannot afford to have high compute power to avoid getting heavier, less power efficient, and costlier. In contrast to many previous methods which rely on multiple type of cues such as camera ego-motion or 2D pose of the human, we show that a carefully designed network model which relies solely on bounding boxes can not only perform  better but also predicts trajectories at a very high rate while being quite low in size of approximately 17 MB. Specifically, we demonstrate that having an auto-encoder in the encoding phase of the past information and a regularizing layer in the end boosts the accuracy of predictions with negligible overhead. We experiment with three first person video datasets: CityWalks, FPL and JAAD. Our simple method trained on CityWalks surpasses the prediction accuracy of state-of-the-art method (STED) while being 9.6x faster on a CPU (STED runs on a GPU). We also demonstrate that our model can transfer zero-shot or after just 15\% fine-tuning to other similar datasets and perform on par with the state-of-the-art methods on such datasets (FPL and DTP). To the best of our knowledge, we are the first to accurately forecast trajectories at a very high prediction rate of 78 trajectories per second on CPU. 
\end{abstract}


\section{Introduction}
\label{sec:introduction}

We address the problem of forecasting human motion in first person monocular videos using neural networks. We are particularly interested in developing a network which can accurately predict future locations of humans in real-time on machines with low compute and memory capability. To this end, we propose a Recurrent Neural Network (RNN) based framework which relies only on detection bounding boxes of tracked moving humans in the scene. 

While such system requirements as mentioned above, i.e. lightweight and simple, are not much relevant from the perspective of current autonomous cars, they are certainly crucial for systems which have either power, weight, size, or monetary constraints. Modern autonomous cars have access to powerful computers with multiple CPUs and GPUs and therefore their system's real-time performance is not affected by the complexity of the network model used.


Take, for example, a social robot \emph{for all} which cannot afford to have high end sensors and heavy compute resources because of monetary and size constraints. In addition to this, it cannot be sweating most of its compute power on just one task (forecasting pedestrian's motion) if it has to be operational for long. Similarly, consider a visual assistance system for people who are visually challenged, again, in addition to it being accessible to all regardless of their economic standards, it has to be small in size so that it can be carried for long and be power efficient for operational longevity. All the aforementioned constraints mean -- less sensors, and low compute and memory resources.

Current state-of-the works \cite{FPL, DTP, MOF, AOL} in this context rely on multiple types of information related to the scene (including the target humans) and/or the camera motion. For e.g. \cite{FPL} relies on ego motion of the camera  and 2D body pose and location of the humans in the scene. Similarly, \cite{DTP} and \cite{MOF} predict the centroids of future bounding boxes by feeding ether only optical flow vectors (within the bounding box) or provide bounding box coordinates in addition to the optical flow, respectively.  Using depth \cite{sfml} \cite{mobile} also helps the problem but getting such depth reliably is difficult.

In this work, we pose the human motion forecasting as a sequence-to-sequence learning \cite{seq2seq} problem and propose to use a network model essentially comprising of RNNs. As the trajectory of humans when perceived in some reference frame can be considered as a sequence of locations in time and space, RNNs are the most natural choice. In this work we use LSTMs \cite{lstm} as RNN units.

\begin{figure}[t]
  \centering
  \vspace{1.7mm}
  \includegraphics[scale=0.21]{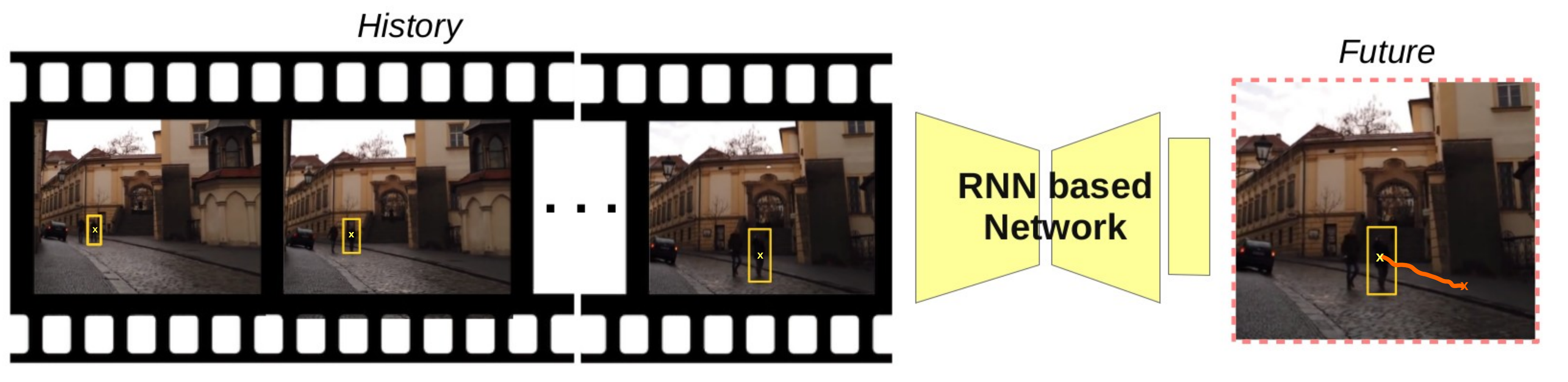}
  \caption{Figure illustrates the task at hand, which is, given past detection bounding boxes of moving humans, we aim to forecast their location in future frames using the proposed RNN based network}
  \label{fig:teaser}
\end{figure}

We design a network model that relies solely on past observations of detection bounding boxes to forecast their future locations (illustrated in Fig. \ref{fig:teaser}). Therefore our model is light-weight to the extent that whole model is only $\sim$ 17 MB in size and runs at 78 Trajectories Per Second (or 4684 fps) on a CPU, considering we predict a trajectory 2 seconds into the future with 1 second input at 30 Hz.

Our model is made up of three components: 1) an encoder block, made up of two LSTMs working as an auto-encoder, that consumes the past bounding box observations and produces a fixed size representation that summarizes the entire history 2) a decoder block which takes the representation vector from previous block to predict the future velocities 3) a trajectory concatenation layer that converts the predicted future velocities into future target locations.  The last layer acts as a regularizer and does not have any parameters to learn (see Sec. \ref{sec:trajConcatLayer} ). The entire network is trained in an end-to-end fashion with two loss terms: auto-encoder loss (Eq. \ref{eq:autoenc_obective}, Sec. \ref{sec:encoderBlock}) and future target location loss ((Eq. \ref{eq:targetObjective}, Sec. \ref{sec:trajConcatLayer})).

The primary motivation for this design choice of having two decoder LSTMs (one in auto-encoder and one for future decoding) was to make sure that network learns to encode the past input observations into a representation that not only extracts all information that is needed to extrapolate the future instances but also to make sure that the learned latent vector (representation) is actually representative of the entire input. This design is inspired from the composite model of LSTM Encoder-Decoder framework proposed in \cite{compositeEncDec}.

We evaluate our model by training it on a recently introduced CityWalks dataset \cite{MOF} which is captured by a person holding a camera and moving in multiple cities with different weather conditions. The network is solely fed with detection bounding box and no other information is provided to it. We show that our trained model not only surpasses the prediction accuracy of the current state-of-the-art \cite{MOF} on the CityWalks datset\cite{MOF}, but also transfers zero-shot to another first person video dataset - First Person Locomotion (FPL) \cite{FPL} which is captured at a different frame rate. To the best of our knowledge none of the previous works have demonstrated a zero-shot transfer to a different dataset, with \cite{MOF} being one exception. However, \cite{MOF} relies on optical flow in addition to the bounding boxes while we rely only on bounding boxes.

We also try to test the same model (trained on CityWalks \cite{MOF}) on a different type of dataset -- Joint Attention for Autonomous Driving (JAAD) \cite{JAAD, JAAD_2}, meant for behavioral understanding of traffic participants. In this dataset. We notice, as expected, that our model does not transfers zero-shot. Interestingly, however, if we fine-tune our trained model on just 15\% of the train sample of JAAD \cite{JAAD} then the performance of our model is on par with the  state-of-the-art \cite{DTP} on this dataset. 

In summary, the contributions of this work are as follows:
\begin{itemize}
    \item We show that a simple but carefully designed/chosen network architecture (Sec. \ref{sec:architecture}) is sufficient to accurately forecast human motion in first person videos (see Sec. \ref{sec:results}) by evaluating it on three datasets: CityWalks\cite{MOF}, FPL\cite{FPL} and JAAD \cite{JAAD, JAAD_2}.
    \item We show that having an extra layer in the end -- Trajectory Concatenation Layer -- with no learnable units improves the over all accuracy of prediction. (\ref{sec:trajConcatLayer})
    \item We also demonstrate that our simple and light-weight model trained on one dataset (CityWalks \cite{MOF}) is capable of transferring zero-shot to another similar dataset ( FPL \cite{FPL}). And when fine tuned with only 15\% train samples of a not so similar dataset (JAAD \cite{JAAD, JAAD_2}), it performs on par with the state-of-the-art \cite{DTP} which was trained on that dataset \cite{JAAD, JAAD_2}.
    \item We show that by the virtue of simplicity of the our network model, we can achieve a real-time performance of predicting $\sim$ 39 Trajectories Per Second (TPS) on a single core of a CPU which is 4.8x faster than our competitor \cite{MOF},  while being only $\sim$ 17 MB in size. On the entire CPU ($>4$ cores), our model can predict 78 TPS i.e. 9.6x faster than the state-of-the-art \cite{MOF}.
    
\end{itemize}

\section{Related Works}

Predicting different aspects of human dynamics has been the focus of computer/robotics vision community for quite some time now. Specifically, in the past decade or half, we have seen remarkable developments in the area of human activity forecasting ( e.g.  \cite{activityECCV,forecastingCORR, peekingInFuture, ng2019forecasting,sun2019relational, rhinehart2017first} ), pose forecasting (e.g. \cite{Mao_2019_ICCV, chiu2019action, pavllo2018quaternet}), and human trajectory forecasting (e.g. \cite{FPL, MOF, AOL, DTP, SSLSTM, socialScene, convTraj, STGAT, locationVelTraj, sophieGAN, convTraj,socialGAN, MXLSTM,MXLSTMPAMI,seeingIsBelieving, relationTraj,karasev2016intent}) - thanks to the incredible developments on CNNs and RNNs. In this work, as we are concerned with the task of predicting motion of humans in the scene, we will constrain our discussion in this section to human trajectory forecasting only.

Most of the works in the context of human trajectory forecasting has been done from the perspective of surveillance, social interaction, crowd behaviour analysis or in sports( e.g.\cite{crowdIJCAI17,crowdICRA17, crowdICRA18, whereWillTheyGo}). Most of these works either rely on the bird's eye view of the scene or depend on multiple camera setup. One of the pioneering work -- SocialLSTM -- introduced in \cite{socialLSTM} used LSTMs to forecast human motion with social pooling. Social pooling was introduced so that network learns the interaction between different pedestrians in the scene while predicting their future. Following it, many similar models were introduced which predict future locations based on social, physical, or semantic constraints. For e.g. SocialGAN \cite{socialGAN} uses a Generative Adversial Network (GAN) framework \cite{GAN} for predicting socially plausible human trajectories, SoPhie \cite{sophieGAN} bases its trajectories on social and physical attention mechanism in addition to GAN in order to produce realistic trajectories, and \cite{socialScene} performs navigation, social, and semantic pooling to predict semantically compliant future trajectories. Similarly, \cite{sceneLSTM} trains both on scene and pedestrian location information to predict future motion. On the other hand,  \cite{STGAT} propose to use temporal correlations of interactions with other pedestrians along with spatial. Few researcher have also considered different information related to the pedestrian, for e.g. \cite{MXLSTM, seeingIsBelieving, MXLSTMPAMI} use a sequence of head poses in addition to other relevant information. For a comprehensive collection of research activities in this area refer \cite{trajPredSurvey}. We are distinct from all the aforementioned works in two ways: 1) we only rely on front view 2) our model depends solely on detection bounding boxes and no other information whatsoever. 

Very recently, we have seen few exciting works in the area of human trajectory forecasting in first person perspective \cite{FPL, MOF, AOL, DTP}. However, all of them rely on multiple information related to the pedestrian whose motion is to be forecasted, scene in which the camera and pedestrians are moving and the ego-motion of the camera. For e.g. \cite{FPL} relies on camera ego-motion and 2D pose of the pedestrians to forecast their motion. Similarly, \cite{MOF, DTP, AOL} all rely on optical flow information and the detection bounding boxes. One slightly different work which forecasts motion of individual skeleton joint locations of the pedestrian was also proposed in \cite{mangalam2020disentangling}. This work too relies on multiple cues such as 2D pose (i.e. skeleton joint information), camera ego motion, and the 3D structure of the scene. In this setting, we are different from all aforementioned works in multiple ways: 1) we only rely on detection bounding boxes of the pedestrians 2) our method can predict a very large number of trajectories per second on CPU, for e.g. our model predicts at 78 TPS which is 9.6x faster than the state-of-the-art on CityWalks \cite{MOF}, 3) we train and test only on CPUs and 4) our network is extremely light weight, it is only $\sim$ 17 MB in size.

\section{Overview}
\label{sec:pipeline}
In this section we start with formulating the problem of forecasting human motion in first person videos and show that this task, by construction, is a sequence-to-sequence problem. Then, we discuss the evaluation metrics common to all the datasets on which we evaluate.

\subsection{Problem formulation}

Consider a scene with a moving human. Assume that the scene is being captured in first person perspective by a freely moving monocular camera. Let us say that at time $t$, we are at frame $f$ in the video sequence. Assume that the human in the scene is being detected and tracked i.e. we have the detection bounding boxes for each frame in the video sequence along with their track ID. The task at hand is as follows: given only the detection bounding boxes of the human in scene for the past $k$ frames   $\{f-k,f-k+1, f-k+2, ..., f\}$, we aim to predict the bounding boxes for the future $f+p$ frames. Formally, consider $B \equiv \{b_{f-k}, b_{f-k+1}, ..., b_{f} \}$  a sequence of bounding boxes in the past $k$ frames relative to the frame $f$ (inclusive of $f$), we want to obtain $P \equiv \{b_{f+1}, b_{f+1}, ..., b_{f+p} \}$ a sequence of bounding boxes for the future $p$ frames. In this work, we use human, person or pedestrian interchangeably.

\subsection{Evaluation Metrics}
We adopt the two commonly used evaluation metric in the trajectory forecasting setting from \cite{socialLSTM}: Average Displacement Error (ADE) and Final Displacement Error (FDE). Average Displacement Error (ADE) is defined as the mean Euclidean distance between the predicted and the ground truth bounding box centroids for all the predicted bounding boxes, and Final Displacement Error (FDE) is defined similarly for the bounding box centroid at the final prediction step only.



\section{Methodology}

\begin{figure*}[!t]
\vspace{6pt}
\begin{minipage}{\linewidth}
    \centering
    \includegraphics[scale=0.3]{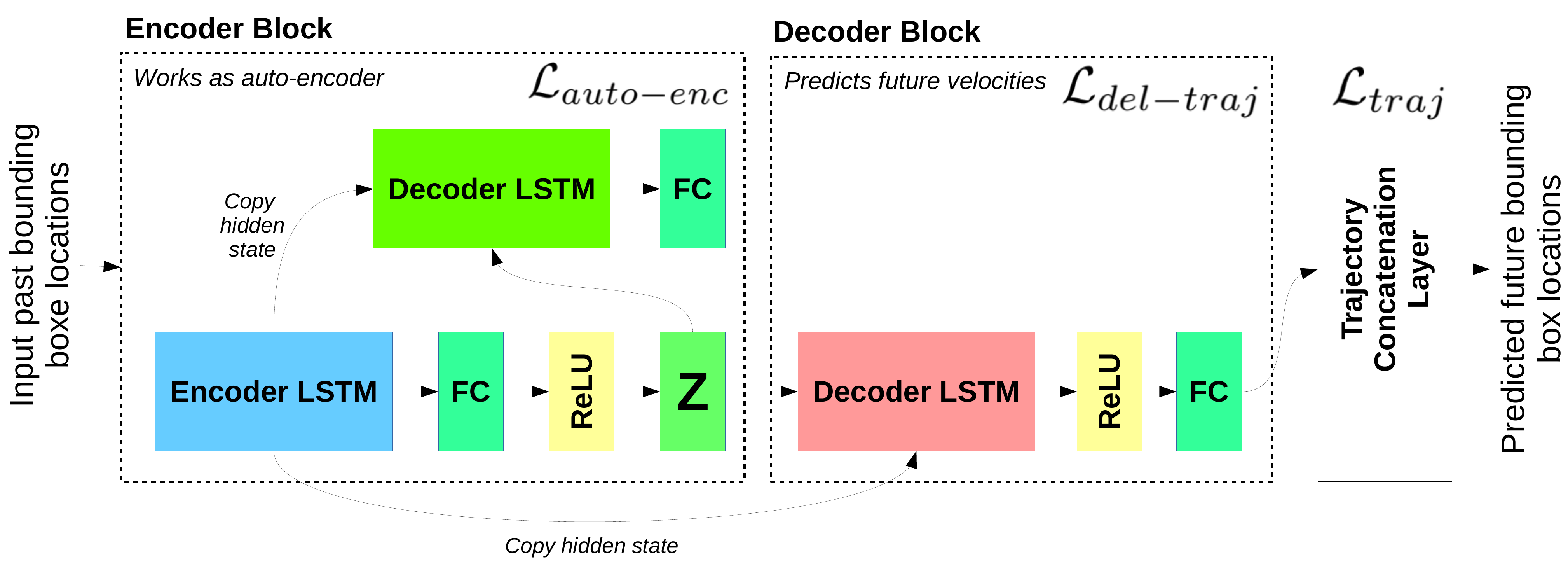}    
    \caption{\textbf{The proposed RNN based network architecture:} The encoder block consumes the bounding box information for the past frames and produces a fixed size representation. The decoder block reads this representation and predicts the velocities of centroid and change in dimension of bounding boxes in future frames. The last layer converts the velocities into locations}
    \label{fig:architecture}
\end{minipage}
\end{figure*}
\subsection{Architecture}
\label{sec:architecture}
Our network model, as shown in Fig. \ref{fig:architecture}, can be understood to be a sequence of three blocks: 1) an encoder block that processes the information for the input sequence and produces a fixed length representation, 2) a decoder block that takes as input the learned representation from the previous block and predicts the information for the future bounding boxes and 3) a trajectory concatenation layer which converts the predicted outputs from the decoder block to actual future locations. The entire model is trained end-to-end and no pre-processing of any sort is performed on the input sequence.

\subsubsection{Encoder Block}  
\label{sec:encoderBlock}
This block of the model is essentially an LSTM based auto-encoder. Let us say that at time $t$ we are at frame $f$ in the video sequence. The input, $\mathcal{I}$, to this block is a sequence of $k$, 8-dimensional vectors, $\mathcal{I} \in \mathbb{R}^{k \times 8}$, which combines different aspects of the human's bounding box information observed over past $k$ frames. Each vector in the sequence is made up of centroid of the bounding box, $ Cxy\in \mathbb{R}^{2}_{+}$, its dimensions in terms of width ($w$) and height($h$), where {$h,w$} $ \in \mathbb{R}_{+}$ , and change in the centroid and its dimension, each in $\mathbb{R}_{+}$. Formally, the input sequence, $\mathcal{I}$, can be written as $ \{B\}_{i=f-k}^{f} \equiv \{ B_{f-k}, B_{f-k+1}, B_{f-k+2}, ..., B_{f}\}$, where each $B$ is an 8-dimensional vector, $B_i = (cx_i, cy_i, w_i, h_i, \Delta{cx}_i, \Delta{cy}_i, \Delta{w}_i, \Delta{h}_i)$ i.e. $B \in \mathbb{R}^{8}$. The $\Delta$ terms (or change terms) are computed as: $\Delta{U}_i = U_i - U_{i-1}$, for $\forall U \in \{cx, cy, w, h\}$.

The Encoder LSTM, $\mathcal{E}_{enc}$, of this block runs through the input sequence, $\mathcal{I} \equiv \{B\}_{i=f-k}^{f}$, and generates a final hidden state vector, $H_{f}^{e}$ which summarizes the complete sequence of bounding box information. The final state vector is then fed to a fully connected layer, $\mathcal{FC}_{enc}^{e}$ which maps it to a vector of $256$ dimensions, $Z_f$. 

The Decoder LSTM, $\mathcal{D}_{enc}$ on the other hand, takes the encoded representation $Z_f$ and runs $k$ times while taking the same $Z_f$ as input at every iteration  to reproduce $k$ hidden states, $\{H_{i}^{d}\}_{i=f}^{f-k}$, one for each iteration, which are then passed through an $\mathcal{FC}_{enc}^{d}$ to map it to the input dimension i.e. $\mathbb{R}^{k \times 8}$. Note that we intentionally reproduce the input sequence in the reverse direction as it has proven to be useful as shown in \cite{seq2seq}. All the above process can be formalized as following (we do not explicitly show the reliance of LSTMs on the hidden states of their previous iterations):

\begin{equation}
    H_{f}^{e} = \mathcal{E}_{enc}(\mathcal{I})
\end{equation}
\begin{equation}
    Z_{f} = \mathcal{FC}_{enc}^{e}(H_f^e)
\end{equation}
\begin{equation}
    \{H_{i}^{d}\}_{i=f}^{f-k} = \mathcal{D}_{enc}(Z_f)
\end{equation}
\begin{equation}
    \hat{\mathcal{I}} = \mathcal{FC}_{enc}^{d}(\{H_{d}^{i}\}_{i=f}^{f-k})
\end{equation}
\noindent\emph{where,} $\hat{\mathcal{I}} \in \mathbb{R}^{k \times 8}$

\noindent\textbf{Why have an auto-encoder?} As the decoder LSTM in this block of the model tries to reproduce the input itself, we can have an objective function (Eq. \ref{eq:autoenc_obective}) which penalizes the model based on how far it is from the actual input. This objective function makes sure that the encoder learns the right representation, in this case $Z_f$, which adequately defines the past information of the bounding boxes. We have empirically shown the effect of this in the Results section (see Table \ref{table:ablation} or Sec. \ref{sec:results}). 

\begin{equation}
    \mathcal{L}_{auto-enc} = \frac{\sum_{i=k-f}^{f}|\hat{\mathcal{I}} \ominus \mathcal{I} |}{k \times 8}
    \label{eq:autoenc_obective}
\end{equation}

\noindent \emph{where}, $\ominus$ represents element-wise vector subtraction operation. There are two things to note here: a) we reverse the input sequence, $\mathcal{I}$, and add negative sign to components corresponding to the velocities and change in dimension of bounding boxes and b) the auto-encoder is not pre-trained and this objective is a part of the overall objective function (Eq. \ref{eq:traj_obective}).

\subsubsection{Decoder Block}
This block comprises of an LSTM $\mathcal{D}_{dec}$, and a FC layer, $\mathcal{FC}_{dec}$, which work in the similar fashion as the decoder of the encoder block with two important differences: a) it runs to predict the future and not to  reproduce the input and b) it only predicts the velocity and dimension change components i.e. we it predicts only the future instances of $\Delta{cx},$ $\Delta{cy}$, $\Delta{w}$ and $\Delta{h}$. 

The working of this block is as follows. It takes the latent representation $Z_f$ from the encoder block and runs $p$ times while taking the same $Z_f$ as the input. At every iteration, it produces a hidden state vector $H_i$, where $i \in \{f+1, f+2, ..., f+p\}$ which is then fed to $\mathcal{FC}_{dec}$,  which maps it to a vector, $V$ of 4-dimension, $( \Delta{cx},$ $\Delta{cy}$, $\Delta{w}$, $\Delta{h})$, i.e. $V \in \mathbb{R}^4_{+}$. Formally, the above process can be defined as follows:

\begin{equation}
    \{H_{i}\}_{i=f+1}^{p} = \mathcal{D}_{dec}(Z_f)
\end{equation}
\begin{equation}
    \{\hat{V_{i}}\}_{i=f+1}^{p} = \mathcal{FC}_{dec}(\{H_{i}\}_{i=f+1}^{p})
\end{equation}

If we choose to apply supervision at this stage of the model then the supervising objective will be the following:

\begin{equation}
    \mathcal{L}_{traj-del} = \frac{\sum_{i=f+1}^{p}|\hat{V} \ominus V|}{p \times 4}
    \label{eq:deltraj_obective}
\end{equation}
\noindent\emph{where}, $V \in \mathbb{R}^{p \times 4}$ is the ground truth for the predicted velocity of the centroid ($\Delta{cx}$, $\Delta{cy}$) and dimension change ($\Delta{w}$, $\Delta{h}$) of the bounding boxes for the future $p$ frames.\\

Every time the decoder LSTM, $\mathcal{D}_{dec}$, starts to decode the future sequences of a trajectory, its hidden state, $H_f$, is always initialized with the final hidden state of the encoder LSTM, $H_f^e$, i.e. $H_f = H_f^{e}$. The motivation for doing this was that as the future motion of the human is not going to be much different from its past motion which is already encoded in the hidden state of the encoder, it makes sense to load the decoder LSTM with this knowledge before it even starts decoding. Informally, consider it as a way of transferring the physics of motion to the future decoder.

\subsubsection{Trajectory concatenation layer}
\label{sec:trajConcatLayer}
This is the last layer of our model. This layer does not predict any new information and acts more like a regularizer. It is composed of a multivariate differentiable function, $\mathcal{G}$, which converts the predicted future velocities of the centroids and the change in dimension of detection bounding boxes into a sequence of locations and dimension of the future bounding boxes.

\begin{equation}
    \{\hat{O}_{i}\}_{i=f+1}^{p} = \mathcal{G}(\{\hat{V_{i}}\}_{i=f+1}^{p},  \mathcal{I}_{f}^{\nabla})
    \label{eq:targetObjective}
\end{equation}

\begin{equation}
  \hat{O}_{f+i} =
    \begin{cases}
      \mathcal{I}_{f}^{\nabla} \oplus \hat{V_{f+i}} &  for, i = 1\\
      \hat{O}_{f+i-1} \oplus \hat{V_{f+i}} & \forall i = 2 ... p
    \end{cases}       
\end{equation}

Where, with slight abuse of notation, $\mathcal{I}_{f}^{\nabla} \in \mathcal{I} $ represents the centroid and dimensions ($w$ and $h$) of the bounding box of the last input frame $f$, i.e. $\mathcal{I}_{f}^{\nabla} = (cx_f, cy_f, w_f, h_f)$, and $\hat{O} \in \mathbb{R}^{p \times 4}$ is the sequence of centroid and dimension information of the predicted future $p$ bounding boxes; $\oplus$ represents element-wise vector addition.

The supervision is applied on the result of this layer. The presence of this layer yields better prediction accuracy (see \ref{sec:ablation}) with negligible overhead as this layer does not have any learnable parameters. The supervising objective function is as follows:

\begin{equation}
    \mathcal{L}_{traj} = \frac{\sum_{i=f+1}^{p}|\hat{O} \ominus O|}{p \times 4}
    \label{eq:traj_obective}
\end{equation}
\noindent\emph{Where}, $O \in \mathbb{R}^{p \times 4}$ is the ground truth centroid ($cx$, $cy$) and dimension($w$, $h$) of bounding box in the predicted sequence of $p$ future frames.

\noindent\textbf{Why does it help?} Supervising on this layer gives us better result because of the fact that this layer which is nothing but a multi-variate differentiable function,  generates multiple new constraints for every predicted vector without adding any extra free parameter to be learned. We believe that this helps the network learn better and hence produce better predictions (see Table \ref{table:ablation}).

\subsection{Training Objective Function}
\label{sec:trainingObjective}
We train our network model end-to-end by minimizing the following objective function which is a linear combination of the above discussed objectives, $\mathcal{L}_{auto-enc}$ and $\mathcal{L}_{traj}$, in Eq. \ref{eq:autoenc_obective} and Eq. \ref{eq:traj_obective}, respectively: 

\begin{equation}
    \mathcal{L} = \alpha \cdot \mathcal{L}_{auto-enc} + \beta \cdot \mathcal{L}_{traj}
    \label{eq:final-obective}
\end{equation}

\noindent\emph{Where}, $\alpha \in \mathbb{R}_{+}$ and $\beta \in \mathbb{R}_{+}$ are hyper-parameters which decide the importance of the corresponding loss term. We have also  conducted an ablation study (see Sec. \ref{sec:ablation}) to evaluate the performance of different objective functions (see Eq. \ref{eq:autoenc_obective}, Eq. \ref{eq:deltraj_obective} and Eq. \ref{eq:traj_obective}) and their combination.


\subsection{Network Implementation Details}

We implemented our network model using PyTorch \cite{pytorch} deep learning framework. All the RNNs were implemented using a single layer LSTM comprising of $512$ hidden units with zero dropout.

The final hidden state of the encoder LSTM of the encoder block is passed through a ReLU unit before being fed into the the FC layer of the encoder block. The FC layer of the encoder block maps its input $512$ dimensional hidden vector into a $256$ dimensional vector. This vector is then read by both the decoder LSTMS following which we have two FCs to map them to corresponding output dimensions: FC coming after the decoder LSTM of the encoder block maps the $512$ hidden vector back to 8-dimensional vector resembling the input where as the FC coming after the decoder LSTM of the decoder block maps the $512$ dimensional hidden state vector into 4-dimensional vector. The 4-dimensional vector is fed to our global trajectory layer to produce actual locations of the future bounding boxes. The auto-encoder is active only during the training and is switched off while testing to save unnecessary time and compute power consumption.

 \section{Results}
\label{sec:results}

\subsection{Datasets}
We evaluate our model on three recently proposed datasets- CityWalks \cite{MOF}, First Person Locomotion (FPL) \cite{FPL} and Joint Attention for Autonomous Driving (JAAD) \cite{JAAD}, which are all captured in first person perspective. While we train only  CityWalks \cite{MOF}, we evaluate on all the three. While CityWalks and FPL datasets are quite similar in nature, JAAD is created for a different purpose, mainly created for the study of behaviour study of traffic participants.

\subsubsection{CityWalks \cite{MOF}}
Citywalks dataset comprises of 358 video sequences  which are captured by a hand held camera in first person perspective. The video sequences are captured in 21 different cities of 10 European countries during varying weather conditions. All the videos are shot in a resolution of $1280\times720$ at 30 Hz frame rate. The dataset also provides two sets of detections: one obtained from YOLOv3 \cite{yolov3} detector and the other acquired using Mask-RCNN\cite{maskrcnn}. In addition to detection bounding boxes, it also provides tracking information. We use only the Mask-RCNN detections for evaluating our work.

\subsubsection{First Person Locomotion (FPL) \cite{FPL}} 
This dataset comprises of multiple video sequences captured by people wearing a chest mounted camera and walking in diverse environment with multiple moving humans. The collective duration of all the video sequences is about $4.5$ hours with approximately $5000$ person observations. The video sequences are captured at 10 Hz (i.e. 10 frames per seconds). This video does not provide detection bounding boxes. However, it does provides the 2D pose of all the humans. As our model relies solely on detection bounding boxes, to use this dataset for our evaluation, we convert the 2D poses into detection bounding boxes. 

\subsubsection{Joint Attention for Autonomous Driving (JAAD) \cite{JAAD}}
This video dataset was primarily created for the study of behaviour of traffic participants. The dataset is made up of $346$ videos with about $82032$ frames captured by a wide-angle camera mounted behind the wind shield below the rear-view mirror of two cars. Most of the videos are captured at a resolution of $1920\times720$ pixels and few are shot at $1280\times720$ pixel resolution. All the video sequences are captured at real-time frame rate i.e. at 30 Hz. This dataset also comes with detection bounding box and track information for each pedestrian in the scene. As the dataset is created with behavioural analysis of traffic participants, it consists of pedestrians involved in different kind of motion behavior for e.g. pedestrians could stop while walking and again start walking, the person reduces or increases speed in the course of motion, etc. and hence this dataset is of not much relevance in our setting. However, we do evaluate our network model on this dataset too by fine-tuning our network (trained on CityWalks \cite{MOF}) on just $15\%$ of its train set.

\subsection{Training details}

We trained our network on the CityWalks dataset \cite{MOF}. Similar to \cite{MOF}, we split the entire dataset into three folds and perform a 3-fold cross validation. At every evaluation two of the 3 folds serve as training and the other as test. We tune the hyper-parameters only on the train fold and test it on the test fold which that particular network has never seen. The reported performance is the average performance of all the three train-test combinations (see Table. \ref{table:evalCitywalk}). 

For training, the bounding box tracks in the train set are split into multiple 90 frames mini-tracks by sliding over each track with a stride of 30 frames. This way we obtain mini-trajectories of 3 second length. We train our model to predict the location and dimension of bounding boxes 2 seconds into the future by observing past 1 second data. In other words, the network is trained to take the bounding box information of past 30 frames and predict the centroid locations in  $60$ future frames. The network is supervised based on the training objective discussed in section Sec. \ref{sec:trainingObjective}. 


The entire network is trainied end-to-end on a CPU (Intel Xeon CPU E5-2650 v4 at 2.20GHz) with 24 cores, without pre-training of any component. We do not use any GPU for training or testing. The network is trained in batches of 200 for 30 epochs with a starting learning rate of 0.00141. The learning rate is halved every 5 epochs. The hyper-parameters $\alpha$ and $\beta$ in Eq. \ref{eq:final-obective} were set to 1.0 and 2.0. The model is optimized using L1 loss using Adam optimizer \cite{adam} with no momentum or weight decay.


\subsection{Performance Evaluation}

\subsubsection{Baseline models}
\begin{itemize}
    \item \textbf{Spation-Temporal Encoder-Decoder (STED)} \cite{MOF}: It is a GRU and CNN based encoder-decoder architecture which relies on bounding box and optical flow information to forecast future bounding boxes. We train on CityWalks \cite{MOF} to compare with this state-of-the-art model (STED \cite{MOF}) on CityWalks.
    \item \textbf{First Person Localization (FPL)}\cite{FPL}: The model introduced in this work relies on 2D pose of pedestrians extracted using OpenPose \cite{openPose} and ego-motion estimates of the camera using \cite{unsupervisedEgoMotion} to predict the future locations of the human. We compare with \cite{FPL} by transferring zero-shot to FPL dataset. One important thing to note is that this dataset is captured at 10 Hz while our model was trained on CityWalks captured at 30 Hz.
    
    \item \textbf{Dynamic Trajectory Prediction (DTP)} \cite{DTP}:
It uses CNN to forecast the future trajectory using past optical flow frames. To compare with DTP \cite{DTP}, we fine-tune our network on just 15\% of training samples of JAAD  dataset\cite{JAAD}.

\end{itemize}

\subsubsection{Results on CityWalks dataset \cite{MOF}}
The performance of our model on CityWalks dataset is presented in Table. \ref{table:evalCitywalk} where we compare with the all models proposed by the current state-of-the-art \cite{MOF} on this dataset. Similar to \cite{MOF}, our model was trained to predict a sequence of bounding box centroids for 60 time steps into the future by observing bounding boxes of past 30 time steps, i.e. we predict 2 seconds into the future; as discussed earlier, in contrast to us \cite{MOF} also takes optical flow as input.

It is clear from Table. \ref{table:evalCitywalk} that that our simple RNN based architecture (trained on Mask-RCNN \cite{maskrcnn} detections) consistently performs better than the STED model \cite{MOF} and all its variants. In the table, we also report how much we improve over the corresponding model in percentage (\%) shown in columns Improvement(ADE) and Improvement(FDE); this metric is computed as: $|dm-om|/dm$, where $dm$ and $om$ are the performances of the other and our models, respectively. While we surpass the prediction metrics for all variants proposed in \cite{MOF}, it is interesting to see our model performing approximately 16\% and 27\% better than BB-encoder variant of \cite{MOF}  as, just like us, this variant does not use optical flow and relies solely on bounding boxes. We believe this performance is mainly due to the the presence of an extra decoder in the encoding phase and the global trajectory concatenation layer.

\begin{table}
\vspace{3.4pt}
\caption{Results on CityWalks datast \cite{MOF}. BB-encoder and OF-encoder of \cite{MOF} take only bounding box and optical flow information, respectively. The term `both' means the results are average of the performance of networks trained with YOLOv3 and Mask-RCNN detections. Improvement columns for the corresponding metric means how much we are better compared to different models. All the models observe past 1s data and predict 2s into the future at a frame rate of 30Hz}
\label{table:evalCitywalk}
\begin{center}
\begin{tabular}{ccccc} \toprule
    {Model}         &  {ADE}    & {FDE} & {Our} & {Our}\\
                    &           &       & Improvement & Improvement \\
                    &           &       &   {(ADE)}    &     {(FDE)}     \\
    \midrule
    STED &   26.0     &  46.9 &   16.88\%     &   4.54\% \\
    (Mask-RCNN) & & & \\
    STED    &   27.4     &  49.8 &   21.13\%     &   10.1\% \\
    (YOLOv3) & & & \\
    BB-encoder     &   29.6     &  53.2 &   27.0\%        &   15.85\% \\
    (both) & & & \\
    OF-encoder     &   27.5     &  50.3 &   21.41\%     &   11.0\%  \\
    (both) & & & \\
    \midrule
    \textbf{Ours} &  \textbf{21.61}    &   \textbf{44.77}  & & \\
    (Mask-RCNN) & & & \\
     \midrule
\end{tabular}
\end{center}
\end{table}

\subsubsection{Zero-shot transfer on FPL dataset \cite{FPL}}
To demonstrate the efficacy of our model which was trained on CityWalks \cite{MOF}, we directly deploy it on the test set of the FPL dataset \cite{FPL} and compare it with the models proposed in \cite{FPL} (see Table \ref{table:fpl}). One important thing to note is that this dataset is captured at 10 Hz while CityWalks \cite{MOF} is captured at 30 Hz.
To evaluate, just like \cite{FPL}, we take a  sequence of boxes from past 10 frames and predict for 10 and 20 future frames. As presented in Table \ref{table:fpl}, we perform better than the constant velocity, Nearest Neighbor, ans Social LSTM \cite{socialLSTM} based methods by a considerable margin (note that these were directly acquired from \cite{FPL} as we test on same the test set provided by \cite{FPL}). Additionally, we also perform better than a  variant (FPL($L_{in}$)) of FPL(Main) model \cite{FPL} which takes only centroid of the pedestrians. We, however, underperform when compared to the main model of \cite{FPL}, (FPL(Main)), which takes 2D pose, centroid location and scale of the pedestrian, and ego-motion of the camera; surprisingly we outperform the FPL(Main) model \cite{FPL} when predicting a longer sequence.

\begin{table}
\vspace{3.4pt}
\caption{Results for zero-shot transfer on FPL dataset \cite{FPL} i.e. our network did not see the FPL dataset while training. FPL(Main) is the primary model of \cite{FPL} that takes camera ego-motion, scale, centroid and pose of the pedestrian as input. The FDE@$t$ means the Euclidean distance between the prediction and ground truth at $t^{th}$ future frame. All the models below observe past $1s$ data (in our case only bounding box) and predict $1s$ and $2s$ into the future as FPL is recorded at 10Hz frame rate.}
\label{table:fpl}
\begin{center}
\begin{tabular}{ccc} \toprule
    {Model}                 &   {FDE@10}    & {FDE@20} \\ \midrule
    ConstVel                 &   107.15      &   -\\
    NNeighbor               &   98.38       &   -\\
    SocialLSTM               &   118.10      &   223.16\\ \midrule
    FPL ($L_{in}$)          &   88.16       &   -   \\ 
    FPL ($X_{in}$)          &   81.86       &   - \\
    FPL (Main)               &   \textbf{77.26}       &   124.42  \\ \midrule
    \textbf{Ours} (zero-shot)        &   85.28       &   \textbf{119.21} \\ 
    \midrule
\end{tabular}
\end{center}
\end{table}

\subsubsection{Results on JAAD dataset \cite{JAAD}}
The primary objective of evaluating our model on this (not so similar) dataset, was to see how well our model handles different kind of behavior based motion of pedestrians. Note that this dataset was created for studying behaviour of traffic participants (we consider only humans in this context). In this datast, humans can be observed moving in  ways which one does not encounter in FPL \cite{FPL} or CityWalks\cite{MOF} datasets. For e.g. we see humans slowing down or stopping after walking some distance, or accelerating after few time steps, etc.. As expected, our model does not directly transfer to this dataset as shown in Table \ref{table:jaad}. However, after fine-tuning our model with just 15\% of the training sample (randomly sampled from sequences 1-250) it performs on par with the state-of-the-art method \cite{DTP} for the test set(sequence 251-346) of this dataset. The method proposed in \cite{DTP} takes optical flow frames to predict future locations of pedestrians. Again, as the test set for us and \cite{DTP} are same, we directly acquire the prediction performance for Constant Acceleration and Constant  Velocity methods from \cite{DTP}.

      
\begin{table}
\vspace{3.4pt}
\caption{Results on JAAD dataset \cite{JAAD} with 15\% fine-tuning. We compare with the DTP model \cite{DTP}.  The FDE@$t$ means the Euclidean distance between the prediction and ground truth at $t^{th}$ future frame. To compare with DTP \cite{DTP}, we down-sampled the frame rate to 15Hz. All the errors are reported with respect to frame size of $1280\times720$ pixels}
\label{table:jaad}
\begin{center}
\begin{tabular}{cccc} \toprule
    {Model}  & {FDE@5} & {FDE@10} & {FDE@15}\\ \midrule
    Constant Acceleration (CA) &  15.3 & 28.3 & 52.8 \\
    Constant Velocity (CV) &  16.0 & 26.4 & 47.5 \\
    \midrule    
    DTP (5 optical flow frames) &  9.4 & \textbf{19.3} & 35.6 \\
    DTP (9 optical flow frames) &  9.2 & 18.7 & 34.6 \\
    \midrule
    \textbf{Ours} (10 Bounding boxes) &  20.39 & 43.88 & 70.41 \\
    (zero-shot transfer) & & & \\
    \midrule
    \textbf{Ours} (6 Bounding boxes)  &  \textbf{9.07} & 19.88 & \textbf{35.52} \\
    (15\% fine-tuning) & & & \\
    \textbf{Ours} (10 Bounding boxes) &  \textbf{7.72} & \textbf{17.84} & \textbf{34.20} \\
    (15\% fine-tuning) & & & \\
    \midrule
\end{tabular}
\end{center}
\end{table}

\subsubsection{Time and memory efficiency}

Our network is capable of forecasting trajectories at a rate of 78 trajectories per second or 4684 fps on CPU (Intel Xeon CPU E5-2650 v4 at 2.20GHz) with more than 4 cores (see Table \ref{table:timeEfficiency}). This is an extremely high rate when compared with the state-of-the-art \cite{MOF} which has a CNN for computing optical that itself takes $123$ms for one frame. In other words, if we ignore the overhead of other components of the STED \cite{MOF}, it still runs at only 8.1 trajectories per seconds meaning we are aprrox. 9.6x faster than STED \cite{MOF} and perform better. At the same time, our model is also extremely light-weight and is of only 17.4 MBs in size.

\begin{table}
\vspace{3.4pt}
\caption{Time efficiency of our model on CPU cores}
\label{table:timeEfficiency}
\begin{center}
\begin{tabular}{ccccc} \toprule
    {CPU}  &   {Trajectories Per Second } & Faster than SOTA \cite{MOF}  & {FPS} \\ 
    (cores)& (TPS) & (TPS) & \\
    
    \midrule
    1       &   38.91 & 4.79x  & 2334\\
    2       &   54.05 & 6.65x   & 3243\\
    4       &   65.87 & 8.10x   & 3952\\
    $>$4    &   78.06 & 9.6x    & 4684\\ 
   
    \midrule
\end{tabular}
\end{center}
\end{table}

\subsection{Ablation Study}
\label{sec:ablation}
In this section we do a thorough ablation study to understand the impact different components of our model on CityWalks \cite{MOF} (see Table \ref{table:ablation} ). Specifically, we train three models 1) $\mathcal{L}_{traj-del}$: with no auto-decoder in the encoder block i.e. without any auto-encoder loss and no global concatenation layer 2) $\mathcal{L}_{traj}$: with global trajectory layer but without the auto-decoder in the encoder block (Sec. \ref{sec:encoderBlock} and 3) Ours ($\mathcal{L}_{traj} + \mathcal{L}_{auto-enc}$): this is our main model comprising of all the components. We take a sequence of bounding boxes for the past 30 time steps and predict for future 15, 30, 45, and 60 frames. Note that we report the results for the best split. We show that each component adds to the performance and reduces the displacement error for all the cases shown in Table \ref{table:ablation}.

\begin{table}
\vspace{3.4pt}
\setlength\tabcolsep{3pt}
\caption{Ablation study of our model on CityWalks dataset \cite{MOF}. $\mathcal{L}_{traj-del}$: means neither auto-encoder nor the trajectory concatenation layers are active, $\mathcal{L}_{traj}$: means we have trajectory concatenation layer but auto-encoder is off and ($\mathcal{L}_{traj} + \mathcal{L}_{auto-enc}$): means both are active. }
\label{table:ablation}
\begin{center}
\begin{tabular}{ccccc} \toprule
    {Input} & {Predicted}  & {$\mathcal{L}_{traj-del}$} &   {$\mathcal{L}_{traj}$}    & {Ours}\\ & & & & ($\mathcal{L}_{traj} + \mathcal{L}_{auto-enc}$) \\
    & & (ADE / FDE) & (ADE / FDE) & (ADE / FDE) \\
    \midrule
    30  & 15    & 6.49 / 11.17 & \textbf{6.44} / 10.97 & 6.46 / \textbf{10.91} \\
    30  & 30    & 11.23 / 20.22    & 10.99 / 19.61 & \textbf{10.93} / \textbf{19.36} \\
    30  & 45    & 16.24 / 31.78  & 15.81 / 30.92 & \textbf{15.71} / \textbf{30.71}\\
    30  & 60    & 21.77 / 44.45   & 21.27 / 44.14 & \textbf{21.11} / \textbf{43.64}\\
    
    \midrule
\end{tabular}
\end{center}
\end{table}

\section{Conclusions}
\label{sec:conclusions}

We presented a simple yet efficient and light-weight RNN based network architecture for predicting motion of humans in first person monocular videos. We discussed how having an auto-encoder in the encoding phase and a regularizing layer in the end helps us get better accuracy. We showed that our method which relied entirely on detection bounding boxes can not only perform better on datasets on which it is trained, but it was capable of transferring zero-shot on a different dataset. We also demonstrated that by fine-tuning on 15\% of the train set of a not so similar dataset, our model is capable of performing on par (even marginally better) with the state-of-the-art method which was trained on this dataset. Also, by the virtue of the simplicity of our network, our model could predict more accurate trajectories at almost 9.6x faster than state-of-the-art while running only on a CPU and yet be of only 17.4 MB in size. 



\bibliography{bibfiles/references}
\bibliographystyle{styles/IEEEtran}


\end{document}